\documentclass[11pt,letterpaper]{article}
\usepackage{emnlp2017}
\usepackage{times}
\usepackage{latexsym}
\usepackage{comment}
\usepackage{multirow}

\emnlpfinalcopy

\def\emnlppaperid{***}

\title{Automated Word Stress Detection in Russian}

\author{Maria Ponomareva {\normalfont and} Kirill Milintsevich {\normalfont and} Ekaterina Chernyak {\normalfont and} Anatoly Starostin\\
National Research University Higher School of Economics, Russian Federation\\
  {\tt maponomareva\_2@edu.hse.ru, knmilintsevich@edu.hse.ru}\\
  {\tt echernyak@hse.ru, anatoli.starostin@gmail.com} \\}

\date{}

\begin{document}

\maketitle

\begin{abstract}
In this study we address the problem of automated word stress detection in Russian using character level models and no part-speech-taggers. We use a simple bidirectional RNN with LSTM nodes and achieve the accuracy of 90\% or higher. We experiment with two training datasets and show that using the data from an annotated corpus is much more efficient than using a dictionary, since it allows us to take into account word frequencies and the morphological context of the word.

\end{abstract}

\section{Introduction}
Character level models and character embeddings have received a lot of attention recently. The character embeddings were used for several NLP tasks, such as word similarity \cite{wieting2016charagram}, sentence similarity \cite{wieting2016charagram}, part-of-speech tagging \cite{wieting2016charagram}, NER \cite{klein2003named}, speech recognition \cite{mikolov2012subword}, question answering \cite{lukovnikov2017neural}, language identification \cite{jaech2016hierarchical}, etc.

In this study we concentrate on a lesser known problem, which to our knowledge has not been completely solved yet, namely the automatic detection of word stress. For some languages, e.g. Russian, this problem might be crucial for speech processing and generation.

Only a few authors touch upon the problem of automated word stress detection in Russian. Among them, one research project in particular is worth mentioning \cite{hall-sproat2013EMNLP}. The authors restricted the task of  stress detection to finding the correct order within an array of stress assumptions where valid stress patterns were closer to the top of the list than the invalid ones. Then, the first stress assumption in the rearranged list was considered to be correct. The authors used the Maximum Entropy Ranking method to address this problem \cite{collins2005russian} and took character bi- and trigrams, suffixes and prefixes of ranked words as features as well as suffixes and prefixes represented in an “abstract” form where most of the vowels and consonants were replaced with their phonetic class labels. The study features the results obtained using the corpus of Russian wordforms generated on the basis of Zaliznyak's Dictionary (approx. 2m wordforms). Testing the model on randomly split train and test samples showed the accuracy of 0.987. According to the authors, they observed such a high accuracy because splitting the sample randomly during testing helped the algorithm  benefit from the lexical information i.e. different wordforms of the same lexical item often share the same stress position. The authors then tried to solve a more complicated problem and tested their solution on a small number of wordforms for which the paradigms were not included the training sample. As a result, the accuracy of 0.839 was achieved. The evaluation technique that the authors proposed is quite far from real-life application which is the main disadvantage of their study. Usually the solutions in the field of automated stress detection are applied to real texts where the frequency distribution of wordforms differs drastically from the one in a bag of words obtained from “unfolding” of all the items in a dictionary.

In addition, another study \cite{reynolds-tyers2015NODALIDA} describes the rule-based method of automated stress detection without the help of machine learning. The authors proposed a system of finite-state automata imitating the rules of Russian stress accentuation and formal grammar that partially solved stress ambiguity by applying syntactical restrictions. Thus, using all the above-mentioned solutions together with wordform frequency information, the authors achieved the accuracy of 0.962 on a relatively small hand-tagged Russian corpus (7689 tokens) that was not found to be generally available. We can treat the proposed method as a baseline for the automated word stress detection problem in Russian.

In many languages, such as French, Czech, Finnish and German the rules for automated word stress detection can be formalized quite easily. Nevertheless there are languages where phonological characteristics do not predict stress position, for instance such word prosodic systems can be found in North-West Caucasian (Abkhaz) and Balto-Slavic languages (Lithuanian, Serbo-Croatian, Russian) \cite{van1999word}.

In Russian every word has one and only one stressed syllable. Lexical stress is free in its positioning (any syllable can be stressed as shown in (1)) and is movable (for many lexemes lexical stress depends on the word form, as shown in (2)).

\begin{enumerate}

\item  \textit{ {\'e}ta} [This-Sg.F.Nom]	 \textit {neyros{\'e}t'} [network-Gg.Nom]	 \textit {b{\'u}det} [be-3Sg.Fut]	 \textit {rasstavl'{\'a}t'} [put-Inf]	 \textit {udar{\'e}niya} [stress-Pl.Acc] 	 \textit {в} [in]	 \textit {slov{\'a}h} [word-Pl.Loc]	 \textit {r{\'u}sskogo} [russian-Sg.M.Gen]	 \textit {yazik{\'a} } [language-Sg.Gen]
							
\item  \textit{d{\'e}revo} [tree-Sg.Nom] \textit{der{\'e}vya} [tree-Pl.Nom]
 
Lexical stress can be crucial in disambiguating between homographs, both between two wordforms ((3)) as well as between two lexemes ((4)):
 
\item  \textit{ruk{\'i}} [hand-Sg.Gen]	 \textit{r{\'u}ki} [hand-Pl.Nom]
 
\item  \textit{b{\'e}regu} [river.bank-Sg.Dat]	 \textit{bereg{\'u}} [protect-1Sg.Pres]

\end{enumerate}

The position of lexical stress in Russian depends on many factors including the morphological content of the word, but also the type of word formation, its frequency and its meaning. A complex system of markers which are defined for all morphemes has been developed in fundamental research \cite{Zaliznyak1985}. There are rules that define the hierarchy and interaction of markers but some of them are not strict and can be considered more of a tendency.
 
For practical purposes the dictionary approach to text accentuation can be appropriate. It is possible to imagine a system that finds an accented form for each token using some predefined list. However such a system would have several disadvantages,  the most important of which would be its inability to predict stress for unknown words.
 
In this paper we propose a formal approach to the problem of automatic accentuation of Russian text by trying to exploit neural character models for these purposes. Furthermore, we try to avoid using any additional or third-party tools for part of speech tagging and try to develop a simplistic approach that is based only on using the training data.

\section{Datasets}
We considered two datasets:
\begin{enumerate}
	\item Zaliznyak's Russian Grammar Dictionary, which lists over 100,000 lexemes \cite{Zaliznyak1985}. Each lexeme and its wordforms are stressed. The dictionary was split into a train and test datasets in a 2:1 ratio, so that all forms of one lexeme belong either to the train or to the test dataset and no lexeme belongs to both. We've assigned the name Dictionary Model (DictM) to the RNN trained on this dataset.
	\item Transcriptions from the speech subcorpus\footnote{ Word stress in spoken texts database in Russian National Corpus [Baza dannykh aktsentologicheskoy razmetki ustnykh tekstov v sostave Natsional’nogo korpusarusskogo yazyka], http://www.ruscorpora.ru/en/search-spoken.html} of Russian National Corpus (RNC) \cite{grishina2003spoken}. The spoken corpus was collected by recording people talking in different situations after which it was transcripted and annotated with word stress, also the transcripts of the Russian movies were included. The main difference between these transcriptions and Zaliznyak's Dictionary is that the transcription usually doesn’t contain all forms of a word and, more importantly contains word contexts, i.e. previous words. By taking context into account we can attempt to differentiate between cases such as \textit{``{\'o}blaka''} [cloud-Sg.Gen] and \textit{``oblak{\'a}''} [clouds-Pl.Nom], since the previous word in most instances will reveal whether the word is singular or plural. This dataset was split into train and test datasets using the same 2:1 ratio.
	We trained two models on the corpus. Let us call the first RNN a Context Dependant Model (CDM). In order to take the previous word into account we used the following algorithm: if the previous word has less than three letters, we remove its word stress and concatenate it with the current word (for example, \textit{``te\_oblak{\'a}''} [that-Pl.Nom cloud-Pl.Nom]). If the previous word has 3 or more letters, we use the last three, since Russian endings are typically 2-3 letters long and derivational morphemes are usually located on the right periphery of the word. As such  we get, for example \textit{``ogo\_{\'o}blaka''} [Sg.N.Gen cloud-Sg.Gen] from \textit{``belogo\_{\'o}blaka''} [white-Sg.N.Gen cloud-Sg.Gen]. The second model (Context Free Model, CFM) has the same architecture but it doesn't take context into account.
\end{enumerate}

\section{Architecture}
We adopted a character level architecture from standard tutorials on Keras framework\footnote{http://keras.io}. Our neural network is a bidirectional recurrent neural network with 64 LSTM nodes and dropout regularization. Every input word is represented by a 40 by 33 matrix, where 40 stands for the maximum observed word length in characters. Shorter words are padded with a padding symbol. 33 is the number of letters in the Russian alphabet and every letter in a word is encoded with one-hot encodings. 

Stress can be considered a characteristic of a vowel that has two possible values. A syllable in Russian has a (C)V(C) structure, so the number of vowels equals the number of syllables and in every word only one vowel will be stressed.  The word stress is encoded by one-hot encoding too and shows which of the 40 letters is annotated with the word stress. The output layer of the RNN again has 40 nodes and is activated by \texttt{softmax}. In order to evaluate the quality of word stress detection we used accuracy.

\section{Results and discussion}
While testing\footnote{Our implementation of the method can be found here: http://github.com/MashaPo/accent\_lstm} the presented approach on Zaliznyak's Dictionary we had 1,767,041 instances in the train dataset and 878,306 instances in the test dataset. We trained DictM for 10 epochs and received the best results on the fourth epoch with 88.7\% accuracy for the test set from the dictionary. The score can be compared with the results of the second experiment in \cite{hall-sproat2013EMNLP} and proves that RNN is as efficient as Maximum Entropy Ranking for this problem.
 
The second dataset was slightly bigger and comprised 2,306,776 unique train instances and 1,154,067 unique test instances. We used this dataset to train CDM and CFM for 10 epochs. CDM achieved the best results during the fifth epoch with 97.7\% accuracy on all words. CFM showed the highest accuracy of 97.9\% during the sixth epoch.
 
The significant difference between those values shows that taking the previous context into account increases the accuracy, although by using corpus we could have ignored some complex cases that are not widely used in actual speech but are present in Zaliznyak's dictionary and increase the weight of most frequent words that do not necessarily have a common type of stress placement. Here we are referring to numerals and frequent adverbs that have their own special type of stress placement. Due to their frequency such cases negatively influenced the accuracy of DictM.
 
We implemented the following method to compare the RNN's. We used those three models to detect word stress in the test set of the corpus. We then computed the number of correct predictions for words of different length and calculated the micro-average of accuracy for every model. It is worth mentioning that the accuracy value for the DictM dropped in comparison to the score obtained from the dictionary test set (88.7\% for the dictionary test set and 75.1\% for the corpus test set). The results for DictM, CFM, CDM are presented in tables 1, 2 and 3 respectively.

\begin{table}
\centering
\small

\begin{tabular}{|p{1cm}|p{2.5cm}|l|}
\hline
{\bf \# of syllables} & {\bf Correct detections, \%} & {\bf Correct detections}\\\hline
2 & 0.690 & 182,285 of 263,952 \\
3 & 0.721 & 127,012 of 176,144 \\
4 & 0.846 & 85,675 of 101,229 \\ 
5 & 0.918 & 42,124 of 45,879 \\ 
6 & 0.952 & 15,241 of 16,009\\
7 & 0.958 & 3,813 of 3,979 \\ 
8 & 0.96  & 744 of 775\\ 
9 & 0.928 & 156 of 168 \\
\hline \hline
\multicolumn{2}{|l|}{\bf Micro-average} & 0.751 \\
\hline
\end{tabular}

\caption{Word length in syllables and the number of correct detections for Dictionary Model}\label{tab:res1}
\end{table}

\begin{table}
\centering
\small

\begin{tabular}{|p{1cm}|p{2.5cm}|l|}
\hline
{\bf \# of syllables} & {\bf Correct detections, \%} & {\bf Correct detections}\\\hline
2 & 0.981 & 259,179 of 263,952 \\
3 & 0.974 & 171,645 of 176,144\\
4 & 0.975 & 98,707 of 101,229\\ 
5 & 0.975 & 44,774 of 45,879\\ 
6 & 0.972 & 15,567 of 16,009\\
7 & 0.950 & 3,782 of 3,979 \\ 
8 & 0.940 & 729 of 775 \\ 
9 & 0.934 & 157 of 168 \\
\hline \hline
\multicolumn{2}{|l|}{\bf Micro-average} & 0.977 \\
\hline
\end{tabular}

\caption{Word length in syllables and the number of correct detections for Context Free Model}\label{tab:res2}
\end{table}

\begin{table}
\centering
\small

\begin{tabular}{|p{1cm}|p{2.5cm}|l|}
\hline
{\bf \# of syllables} & {\bf Correct detections, \%} & {\bf Correct detections}\\\hline
2 & 0.983 & 259,656 of 263,952 \\
3 & 0.977 & 172,164 of 176,144\\
4 & 0.976 & 98,887 of 101,229\\ 
5 & 0.977 & 44,837 of 45,879\\ 
6 & 0.973 & 15,591 of 16,009\\
7 & 0.955 & 3,802 of 3,979 \\ 
8 & 0.923 & 716 of 775 \\ 
9 & 0.952 & 160 of 168 \\
\hline \hline
\multicolumn{2}{|l|}{\bf Micro-average} & 0.979 \\
\hline
\end{tabular}

\caption{Word length in syllables and the number of correct detections for Context Dependant Model}\label{tab:res3}
\end{table}

\begin{table}
\centering
\small

\begin{tabular}{|p{1cm}|p{2.5cm}|l|}
\hline
{\bf \# of syllables} & {\bf Correct detections, \%} & {\bf Correct detections}\\\hline
2 & 0.756 & 17,852 of 23,606 \\
3 & 0.829 & 5,402 of 6,510\\
4 & 0.823 & 1,011 of 1,227\\ 
\hline \hline
\multicolumn{2}{|l|}{\bf Micro-average} & 0.77 \\
\hline
\end{tabular}

\caption{{\bf CFM} score on 50 homograph pairs}\label{tab:res4}
\end{table}

\begin{table}
\centering
\small

\begin{tabular}{|p{1cm}|p{2.5cm}|l|}
\hline
{\bf \# of syllables} & {\bf Correct detections, \%} & {\bf Correct detections}\\\hline
2 & 0.810 & 19,143 of 23,606 \\
3 & 0.844 & 5,498 of 6,510\\
4 & 0.847 & 1,040 of 1,227\\ 
\hline \hline
\multicolumn{2}{|l|}{\bf Micro-average} & 0.819 \\
\hline
\end{tabular}

\caption{{\bf CDM} score on 50 homograph pairs}\label{tab:res5}
\end{table}

\begin{table}
\centering
\small

\begin{tabular}{|l|l|p{0.6cm}|p{0.3cm}|p{0.6cm}|p{0.3cm}|}
\hline
{\bf Stressed wordform}                & {\bf Total}                 & \multicolumn{2}{p{1.5cm}|}{{\bf CDM accuracy}} & \multicolumn{2}{p{1.5cm}|}{{\bf CFM accuracy}} \\ \hline
\textit{slov{\'a}} [word-Pl.Nom] & 984                   & 0.871                  & \multirow{2}{*}{0.80}  & 1.0                       & \multirow{2}{*}{0.54}    \\
\textit{sl{\'o}va} [word-Sg.Gen] & 812                   & 0.714                  &                        & 0.0                       & \\ \hline
\textit{del{\'a}} [affair-Pl.Nom] & 976                   & 0.929                  & \multirow{2}{*}{0.86}  & 1.0                       & \multirow{2}{*}{0.62}    \\
\textit{d{\'e}la} [affair-Sg.Gen] & 588                   & 0.753                  &                        & 0.0                       &                          \\ \hline
\textit{n{\'o}gi} [leg-Pl.Nom] & 542                   & 0.797                  & \multirow{2}{*}{0.74}  & 1.0                       & \multirow{2}{*}{0.85}    \\
\textit{nog{\'i}} [leg-Sg.Gen] & 92                   & 0.44                  &                        & 0.0                       &                          \\ \hline
\textit{v{\'o}lny} [wave-Pl.Nom] & 88                   & 0.72                  & \multirow{2}{*}{0.77}  & 1.0                       & \multirow{2}{*}{0.60}    \\
\textit{voln{\'y}} [wave-Sg.Gen] & 57                   & 0.85                  &                        & 0.0                       &                          \\ \hline
\end{tabular}

\caption{{\bf CDM} and {\bf CFM} detailed results for some of the homograph pairs}\label{tab:res6}
\end{table}

The comparison of the results proves that training the model using the corpus gives us a visible increase in accuracy even when the left context is not considered. For DictM there is a clear positive correlation between the number of words and the accuracy of predictions, DictM gets better results then CDM on 8- and 9-syllable words, which are rare in the corpus and can be “new” for CFM and CDM, while DictM could have “learned” the whole paradigm. CFM and DFM show negative correlation between the accuracy and the number of syllables which is expected  due to lower frequency of longer words.

Next, the results from CDM clearly present the advantages of training the RNN while taking the previous word into account, since it increases the number of correctly detected word stresses including homograph cases. The similar way of model testing makes our results comparable with those obtained in \cite{reynolds-tyers2015NODALIDA}, our Context Free Model and Context Dependant Model showed higher micro-average of accuracy than the baseline. 

In order to show CDM to be more accurate than CFM due to the homograph disambiguation we conducted additional tests to learn how both models treat the homographs. More precisely, we extracted  the tuples of words from the dictionary that only differed in stress position (\textit{``dorog{\'o}y''} [expensive-M.Sg.Nom] \textit{``dor{\'o}goy''} [road-Sg.Instr] ). Next, we selected such homograph pairs that for both words the number of occurrences in the corpus was above the predetermined threshold. Tables 4 and 5 show the scores after testing CFM and CDM  on 50 most frequent pairs.
More detailed results for four homograph pairs are displayed in Table 6. The data clearly indicates that  CFM simply chooses the most frequent word in a homograph pair. even though CDM makes mistakes when analysing more frequent words in the pair, it significantly increases the accuracy for less frequent words. The overall accuracy for cases where the frequency of the homographs is comparable (rows 1,2 and 4 of the table) is notably higher for CDM than CFM. 

We have also conducted error analysis. First of all, a huge source of errors are proper names both first names and surnames. Several typical Russian surnames are derived from nouns or adjectives and differ from other wordforms only in stress position. We may address this issue by exploiting NER algorithms and introducing special rules for proper names. Another kind of error is related to words with ambiguous word stress. For example, in words like \textit{``musoroprovod''} [garbage.chute-Sg.Nom] two word stress positions are possible in modern Russian: \textit{musoropr{\'o}vod} or \textit{musoroprov{\'o}d}. Last but not the least, in Russian the letter \textit{{\"e}} is always stressed, but if this letter is written as a regular \textit{e}, the RNN may erroneously ignore it.

\section{Future work}
There are a few directions for future work:
\begin{enumerate}
	\item improving the way we take word context into account. We may use more sophisticated techniques to define the ending and morphological features of the previous word. We may also explore how considering the next word improves the performance.
	\item introducing rules for named entities in general and proper names in particular;
	\item experimenting with reducing the number of instances in a train dataset to both lower the training time and to find specific important examples for training;
	\item experimenting with RNNs carefully in order to gain more linguistic intuition on how word stress is chosen.
\end{enumerate}

\section{Conclusions}

In this study we conducted a few experiments on training RNNs to detect word stress in Russian words. Our results show that, first of all, the character level RNNs are quite suitable for the task, since on average we achieve the accuracy around 90\% or higher. Secondly, we explored two different sources of training data (namely, a dictionary and an annotated corpus) and we can definitively state that using the corpus suits the task better, since it allows us to take frequent cases and morphological context into account and use this information for further disambiguation.

\section*{Acknowledgements} The research was prepared within the framework of the Basic Research Program at the National Research University Higher School of Economics (HSE) and supported within the framework of a subsidy by the Russian Academic Excellence Project “5-100”.

\bibliography{sclem}
\bibliographystyle{emnlp_natbib}

\end{document}